\documentclass[nohyperref]{article}

\usepackage{microtype}
\usepackage{graphicx}
\usepackage{booktabs} 

\usepackage{hyperref}

\usepackage[accepted]{icml2022}

\usepackage{amsmath}
\usepackage{amssymb}
\usepackage{mathtools}
\usepackage{amsthm}

\usepackage{wrapfig}
\usepackage{lipsum}

\usepackage{bbm}
\usepackage{amssymb}
\usepackage{amsmath}
\usepackage{enumitem}
\usepackage{graphicx}
\usepackage{mathtools}
\usepackage{xcolor, soul, colortbl}
\usepackage{cuted}
\usepackage{capt-of}
\usepackage{subfig}
\usepackage{natbib}
\usepackage[font=small]{caption}

\definecolor{color1}{RGB}{100, 140, 100}

\newsavebox\MBox

 \def\mathunderline#1#2{\color{#1}\underline{{\vphantom{\mathstrut}\color{black}#2}}\color{black}}
 
\definecolor{l}{RGB}{245,220,215}

\usepackage[capitalize,noabbrev]{cleveref}

\theoremstyle{plain}

\theoremstyle{definition}

\theoremstyle{remark}

\usepackage[textsize=tiny]{todonotes}

\icmltitlerunning{Partial Graph Reasoning for Neural Network Regularization}

\begin{document}

\twocolumn[
\icmltitle{Partial Graph Reasoning for Neural Network Regularization}



\icmlsetsymbol{equal}{*}

\begin{icmlauthorlist}
\icmlauthor{Tiange Xiang}{usyd}
\icmlauthor{Chaoyi Zhang}{usyd}
\icmlauthor{Yang Song}{unsw}
\icmlauthor{Siqi Liu}{paige}
\icmlauthor{Hongliang Yuan}{tencent}
\icmlauthor{Weidong Cai}{usyd}
\end{icmlauthorlist}

\icmlaffiliation{usyd}{University of Sydney}
\icmlaffiliation{unsw}{University of New South Wales}
\icmlaffiliation{paige}{Paige AI}
\icmlaffiliation{tencent}{Tencent AI Lab}

\icmlcorrespondingauthor{Tiange Xiang}{txia7609@uni.sydney.edu.au}
\icmlcorrespondingauthor{Weidong Cai}{tom.cai@sydney.edu.au}

\icmlkeywords{Machine Learning, ICML}

\vskip 0.3in
]



\printAffiliationsAndNotice{}  

\begin{abstract}
	Regularizers help deep neural networks prevent feature co-adaptations. Dropout, as a commonly used regularization technique, stochastically disables neuron activations during network optimization. However, such complete feature disposal can affect the feature representation and network understanding. Toward better descriptions of latent representations, we present \textit{DropGraph} that learns a regularization function by constructing a stand-alone graph from the backbone features. DropGraph first samples stochastic spatial feature vectors and then incorporates graph reasoning methods to generate feature map distortions. This add-on graph regularizes the network during training and can be completely skipped during inference. We provide intuitions on the linkage between graph reasoning and Dropout with further discussions on how partial graph reasoning method reduces feature correlations. To this end, we extensively study the modeling of graph vertex dependencies and the utilization of the graph for distorting backbone feature maps. DropGraph was validated on 4 tasks with a total of 8 different datasets. The experimental results show that our method outperforms other state-of-the-art regularizers while leaving the base model structure unmodified during inference.
\end{abstract}
\section{Introduction}
	
	Dropout \cite{srivastava2014dropout} has been widely used for reducing feature co-adaptation in deep neural networks. With the success of using Dropout as a regularization technique, many recent studies have studied the impact of allowing dropout in neural networks theoretically \cite{baldi2013understanding,wager2013dropout} and empirically \cite{simonyan2014very,krizhevsky2012imagenet}. In a classic paradigm, neurons are dropped randomly following the Bernoulli distribution with a specific dropout rate $\rho$ during training, such that:
	\begin{equation} \label{eq1}
	f_{Dropout}(\mathbf{X}) = \delta\cdot \mathbf{X},
	\end{equation}
	where $\mathbf{X}$ is the incoming signals and $\delta\sim\textit{Bernoulli}(\rho)$ is a gating 0-1 Bernoulli variable, with probability $\rho$ for $\delta$ to be 0 and thus dropping out neuron activations. However, this classic approach is primarily targeted on fully-connected linear layers and has little effect on multi-dimensional feature maps with strong spatial correlations.
	
	Considering spatial tensors in image-oriented Convolutional Neural Networks (CNNs), Tompson \textit{et al}. \cite{tompson2015efficient} experimented with dropping out the feature vectors across all channels, which they called SpatialDropout. When neuron activations pass through such layers, feature vectors are binary gated with only two possible states: identity or none. Several recent works \cite{larsson2017ultra,ghiasi2018dropblock,chen2020dropcluster,tang2020beyond,zoph2018learning} followed the same intuition and applied dropout in more structured forms.
	
	We argue that the gist of Dropout lies in distorting information propagation of co-adapted signals by random weakening inter-neuron dependencies. The neuron activation distribution varies during network optimization and ends up with information spreading over all neurons uniformly. To this end, applying appropriate distortions to the feature maps could accomplish the same regularization effect. This has also been pointed out by \cite{tang2020beyond}. Instead of zeroing out $\mathbf{X}$ directly, signals can also be obscured by binary gating an extra distortion term, which can be formulated as:
	\begin{equation} \label{eq2}
	f_{reg}(\mathbf{X}) = \mathunderline{blue}{\delta\cdot\mathbf{X}} + \mathunderline{red}{(1-\delta)\cdot\mathbf{R}},
	\end{equation} 
	where $\mathbf{R}$ is the appended distortions and $\delta\sim\textit{Bernoulli}(\rho)$. This equation splits Eq. \ref{eq1} into two parts: \textcolor{blue}{the origin term} and \textcolor{red}{the distortion term}, which also generalizes dropout-based methods \cite{srivastava2014dropout,tompson2015efficient} when there is no distortions i.e. $\mathbf{R}=0$.
	
	
	
	In this work, we propose to learn the distortions $\mathbf{R}$ via \textit{partial graph reasoning}. While graph reasoning has been commonly employed to enhance feature representation learning in the computer vision domain \cite{chen2019graph,li2018beyond}, we reformulate it as a Dropout-like regularizer in deep neural networks. Specifically, we build a \textit{stand-alone} graph network that can be inserted into the backbone to learn $\mathbf{R}$ from an input-specific perspective. Graph reasoning with partially sampled feature vectors is designed with residual-based graph convolutions \cite{kipf2016semi} applied to introduce regularizations similar to Eq. \ref{eq2}. With only the most basic operators, our partial graph reasoning module brings little computational overhead during training. During inference, being a stand-alone graph network, we could completely \textit{Drop} the \textit{Graph}, and provide performance improvements to the regularized backbone network without any additional computations.
	
	
	
	
	Our main contributions are two-fold: \textbf{(1)} We propose the very first learning-based regularization framework, namely DropGraph, which dynamically regularizes networks based on partial graph reasoning. A stand-alone graph neural network generates distortions based on randomly sampled feature vectors, and is skipped during inference yet the regularized learning provides marked performance improvement. Motivations and intuitions are clearly clarified with preliminary studies and in-depth discussions. \textbf{(2)} The effectiveness of DropGraph is empirically validated on numerous benchmarks, including: image classification, image semantic segmentation, point cloud classification, and graph recognition. DropGraph outperforms its state-of-the-art counterparts numerically and statistically. Detailed ablative experiments were additionally conducted to provide a comprehensive analysis of our proposed method.

	\section{Related Work}

	\noindent
	\textbf{Dropout-based Network Regularization.} Regularization techniques \cite{huang2016deep, wan2013regularization, ioffe2015batch} in neural networks aim to reduce feature co-adaptation and, therefore, overcome data over-fitting. As one of the most commonly used regularization techniques, Dropout \cite{srivastava2014dropout} randomly wipes out neuron information during network training to penalize model complexity. Recent works have studied the feasibility of dropping out different signal types and proposed several structured dropout variants. DropPath \cite{larsson2017ultra} and ScheduledDropPath \cite{zoph2018learning} drop layer-to-layer connections in multi-branch building blocks. DropBlock \cite{ghiasi2018dropblock,dai2019batch}, as a close extension to SpatialDropout \cite{tompson2015efficient}, drops a block of spatial feature vectors simultaneously. Following DropBlock, AutoDropout \cite{49960} utilized an external controller to search for the most suitable Dropout patterns for each individual architecture. In a recent work, R-Drop \cite{wu2021r} proposed a new training strategy to ensure dropout consistency of sub models in transformer \cite{vaswani2017attention} architectures. Rather than wiping out neuron activations completely, Tang \textit{et al}. \cite{tang2020beyond} adopted a similar approach to ours that generates feature map distortions. However, their so-called distortion policy starts from reducing the intermediate Rademacher complexity, while we focus on modeling feature sample correlations through graph reasoning methods. Moreover, unlike most of these existing approaches, which were mainly evaluated on image classification, our DropGraph is verified on a variety of down-streaming tasks with different data modalities.

\noindent
\textbf{Graph Reasoning.}  Graph structures are considered to better represent inter-relations of signal embeddings than grid-based data structures. Graph Neural Network (GNN) \cite{scarselli2008graph} was first proposed to resolve a graph by updating node embeddings iteratively. Following GNN, Kipf \textit{et al}. \cite{kipf2016semi} presented GCN that adopts learnable kernels to apply convolutions on graphs. During recent years, subsequent studies have been made on learning more effective message passing rules \cite{hamilton2017inductive}, incorporating attention \cite{velivckovic2017graph}, and increasing network depth \cite{li2019deepgcns}. There are also prior works that combine graph convolution modules, i.e., graph reasoning, with CNNs for assisting vision tasks. For example, the two concurrent works \cite{li2018beyond,chen2019graph} have studied the feasibility of learning graph representations from features in 2D-grids. They first fully transform the 2D-grid features into discrete graph embeddings and then reason global information through GCN layers. In this work, we reformulate Dropout-based regularizers as partial graph reasoning to regularizer the backbone network. In contrast to other graph reasoning approaches, our proposed module is only activated during training and can be completely skipped during inference. 


	\section{Methods}
	Consider a graph $\mathbf{G}=(\mathbf{V}, \mathbf{A})$ with a vertex set $\mathbf{V}$ sampled from the backbone feature maps and an adjacency matrix $\mathbf{A}$. We first describe the commonly adopted graph reasoning methods in CNNs and then discuss how such graphs can be used for regularization. Subsequently, we introduce \textit{DropGraph}, a graph-based regularizer that learns distortions. 

	\subsection{Graph Reasoning as Regularization} \label{gr}
	
	
	\textbf{Graph reasoning on a partial graph.} In a standard graph reasoning paradigm, image-space feature maps $\mathbf{X}$ at a layer are first projected into discrete graph node embeddings \cite{li2018beyond,chen2019graph}. A stand-alone graph network consisting of consecutive GCN layers is then built to reason the transformed features and re-project them back to the original feature maps in a residual style, such that:
	\begin{equation} \label{GR}
	   f_{GCN}(\mathbf{X}) = \mathbf{X} + \mathbf{A}\mathbf{X}\mathbf{W},
	\end{equation}
	where $\mathbf{A}$ is the adjacency matrix, and $\mathbf{W}$ is the learnable weights of one GCN layer for simplicity. 
	
	We claim that by constructing $\mathbf{V}$ with only a subset of the feature vectors in $\mathbf{X}$, applying graph reasoning on such \textit{partial graph} still learns meaningful representations and can be re-formulated as a backbone regularizer with further modifications (later in this section). A sampling ratio $\alpha$ can be set up to control the portion of feature vectors in $\mathbf{X}$ to be selected as the graph vertices such that $\mathbf{V}\subseteq\mathbf{X}$ ($\mathbf{V}\equiv\mathbf{X}$ when $\alpha=100\%$ and $\mathbf{V}\equiv\emptyset$ when $\alpha=0\%$). Such partial graph reasoning can be extended from Eq. \ref{GR} as:
	\begin{equation} \label{PGR}
	    f_{PartialGCN}(\mathbf{X}) = (\mathbf{X}\setminus\mathbf{V}) \cup (\mathbf{A}\mathbf{V}\mathbf{W}),
	\end{equation}
	
	\noindent
	where $\mathbf{X}\setminus\mathbf{V}$ represents the complement set of $\mathbf{V}$ in the super set $\mathbf{X}$.
	
\begin{figure*}
    \centering
    \includegraphics[width=0.9\linewidth]{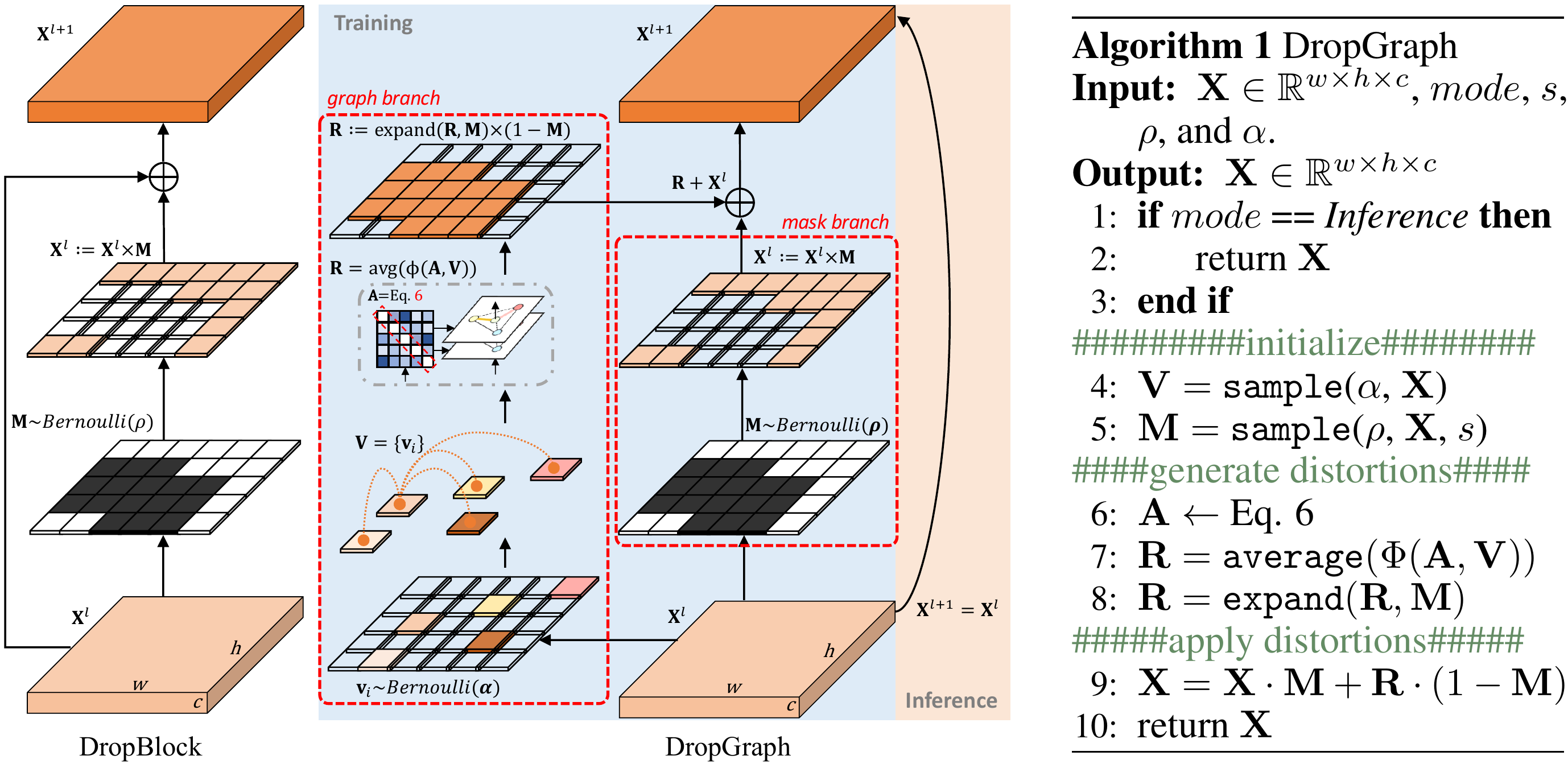}
    \caption{DropGraph learns feature map distortions during network training and can be completely skipped during inference. $w, h, c, s$ denote the feature map width, height, \#channel, and the block size; \texttt{sample} denotes the Bernoulli sampling; \texttt{average} denotes the average pooling operation; \texttt{expand} denotes the expansion of the pooled distortions to all sampled feature vectors for size consistency; ‘$\cdot$’ denotes broadcastly multiplication.}
    \label{component}
\end{figure*}
	
	\textbf{Can partial graph reasoning help network learn better?} To validate whether the stand-alone graph is able to improve the overall performance of backbone network under our partial reasoning setting, preliminary trials are conducted by incorporating Eq. \ref{PGR} on a ResNet-50 network for image classification. In Table \ref{preliminary}, we investigate constructing $\mathbf{V}$ from the feature maps following two heuristics: \textit{(i)} sampling feature tensors with the \textit{top} absolute activations; \textit{(ii)} sampling \textit{random} feature tensors. The impact of different sampling strategies along with sampling ratio $\alpha$ is quantitatively analyzed by either enabling Eq. \ref{PGR} during both network training \& inference or training only. The preliminary experiments offer us the following insights: \textit{(i)} The additional graph reasoning module with sampling ratio $\alpha<100\%$ improves backbone network performance generally (highlighted cells), therefore validating the effectiveness of our formulation in Eq. \ref{PGR}. \textit{(ii)} Enabling graph reasoning modules during training and disabling such modules during inference enhances performance the most.
	
\begin{table}[h]
        \centering
	\caption{CIFAR-100 validation accuracy on sampling the top activated features and random features with different sampling rate $\alpha$. `Train\&Inf.' denotes enabling the graph during both network training and inference while `Train' denotes during network training only. Highlighted cells represent results better than the plain backbone (with $\alpha=0\%$).}
		\begin{tabular}{{l|c c|c c}}
			\toprule
			          & \multicolumn{2}{c|}{Top (\%)} & \multicolumn{2}{c}{Random (\%)}\\
			 $\alpha$ & Train\&Inf. & Train & Train\&Inf. & Train\\
			\hline
			\hline
			$0\%$ & 77.85 & 77.85 & 77.85 & 77.85\\
			\hline
			$12.5\%$ & \cellcolor{l}78.61 & 77.65 & 77.80 & \cellcolor{l}79.29\\
			$25.0\%$ & \cellcolor{l}78.88 & 77.28 & \cellcolor{l}77.92 & \cellcolor{l}79.42 \\
			$50.0\%$ & \cellcolor{l}77.95 & 76.26 & 77.75 & \cellcolor{l}79.03 \\
			$100.0\%$& 75.99 & 75.49 & 75.44 & 2.5\\
			\bottomrule
		\end{tabular}
	\label{preliminary}
\end{table}
	
	
	
	\textbf{Linking between graph reasoning and Dropout.} Following the above insights, we suggest that by disabling graph reasoning during network inference, the partial graph reasoning resembles Dropout (Eq. \ref{eq1}). In this way, we can easily reformulate Eq. \ref{PGR} to the same form as Eq. \ref{eq2} by specifying the distortion term $\mathbf{R}$ as the output of graph reasoning:
	\begin{equation} \label{eq4}
	    f_{reg}(\mathbf{X}) = (\mathunderline{blue}{\delta\cdot\mathbf{X}\setminus\mathbf{V}}) \cup (\mathunderline{red}{(1-\delta)\cdot\mathbf{A}\mathbf{V}\mathbf{W}}),
	\end{equation}
		where $\mathbf{V}\sim\textit{Bernoulli}(\mathbf{X}, \alpha)$, $\delta\sim\textit{Bernoulli}(\rho)$, and $\cdot$ denotes the broadcast pointwise multiplication. By skipping the graph during network inference, the behaviour of such partial graph reasoning aligns to Dropout, and it therefore can be reformulated as another form of regularization. However, questions still remain on how best to define a valid adjacency matrix $\mathbf{A}$ and how to ensure the regularization effect, which will be discussed in the next section.   
	
	\textbf{Why partial graph reasoning regularizes networks?} In CNNs, feature maps are generated by sliding fixed convolution kernels across each pixel with limited receptive fields. Such paradigm enforces local correlations in the same neighborhood and easily leads to feature co-adaptations. Eq. \ref{eq4} alleviates the co-adaptation from two perspectives: first, appended distortions break the conventional convolution pattern and disturb inter-pixel correlations at random positions, which has also been discussed in \cite{tang2020beyond}; second, local pixel responses are replaced by the enhanced graph vertices, enabling long-range dependencies to be captured out of the convolution neighborhood.

	\subsection{DropGraph} \label{dropgraph}
	After obtaining the derivation in Eq. \ref{eq4}, we here introduce a functional module, namely DropGraph that instantiates such re-formulation in the previous section with a graph reasoning architecture, to be equipped in deep neural networks for feature map regularization. DropGraph is a graph-based regularizer that learns to generate distortions from data. During training, the regularization process of DropGraph (Figure \ref{component} left) is achieved via a \textit{mask branch} and a \textit{graph branch}, which can be controlled by the sampling probability $\rho$ and $\alpha$, respectively. The DropGraph algorithm is presented in Figure \ref{component} right.
	
	In the mask branch, we follow Eq. \ref{eq4} to sample the binary gating variables $\delta$ for each feature vector in the incoming feature map $\mathbf{X}$. Instead of gating individual pixels \cite{srivastava2014dropout}, we sample $\delta$ as spatially contiguous square blocks \cite{ghiasi2018dropblock, tang2020beyond}. Therefore, we refer such block-wise gates for the entire feature map as the \textit{dropout mask} $\mathbf{M}$ that is controlled by the block size $s$ and the sampling probability $\rho$.

	
	In the graph branch, we build the graph $\mathbf{G}=(\mathbf{V}, \mathbf{A})$ from the incoming feature map $\mathbf{X}$ and apply partial graph reasoning to generate the distortions as described in Sec. \ref{gr}. Specifically, we construct the graph vertices $\mathbf{V}$ by sampling \textit{random} tensor vectors in $\mathbf{X}$ based on the probability $\alpha$. Then, the adjacency matrix $\mathbf{A}$ is constructed among all vertices $\{\mathbf{v}\}\in\mathbf{V}$ through a similarity measurement function \texttt{sim}$(\cdot, \cdot)$. $\mathbf{A}$ and $\mathbf{V}$ are subsequently fed into a distortion generator $\Phi$ to infer vertex-to-vertex distortions, which are then channel-wise average pooled, expanded and applied on the $\mathbf{M}$ masked input features. During network training, the graph weights $\mathbf{W}$ are jointly optimized with the backbone network by sharing the same gradient flows. To maintain the completeness of feature representations at an initial learning stage, $\rho$ is adjusted from $0\%$ to the target value following a predefined scheduler \cite{zoph2018learning}.

	\textbf{Dependency modeling.} Recall that the purpose of regularizers lies in reducing pixel-to-pixel dependencies. There are two kinds of pixel relationships modeled by the adjacency matrix $\mathbf{A}$ in our graph: intra-dependency (modeled as diagonal values of $\mathbf{A}$) and inter-dependency (modeled as non-diagonal values of $\mathbf{A}$). When $\mathbf{A}$ degrades to an all-zero matrix without modeling any dependencies, DropGraph instantiates to DropBlock. 
	
	
	To activate graph reasoning, dependencies have to be embedded in $\mathbf{A}$. In CNNs, two adjacent feature vectors share similar distributions and hence are very likely to correlate with each other. To ease such adjacency-based correlations, strong connections are expected to be built for any pair of vertices that are apart from each other in the semantic feature space. Specifically, we utilize pair-wise dot product similarity $\texttt{sim}(\cdot,\cdot)$ for constructing $\mathbf{A}$. Semantically closer vertices are assigned with lower values in $\mathbf{A}$ while the dissimilar ones are allocated with higher values. To achieve this, we minus the softmax gated similarity scores from an all-one matrix and then normalize $\mathbf{A}\in(0,1)$ by scaling down with their cardinality:
	\begin{equation} \label{eq5}
	\mathbf{A} = \frac{1 - \texttt{softmax}(\texttt{sim}(\mathbf{v}_i,\mathbf{v}_j))}{\texttt{max}(||\mathbf{V}||-1, 1)},
	\end{equation}
	
	\noindent
	where $||\cdot||$ denotes cardinality and all arithmetic operations above are in vectorized form.
	
	
	\textbf{Distortion generation.} Given the sampled vertex set $\mathbf{V}$ and the adjacency matrix $\mathbf{A}$, we utilize a set of basic graph message passing rules $\Phi(\mathbf{V},\mathbf{A})$ to generate distortions similarly to Eq. \ref{PGR}, but in a residual style. Towards better efficiency and less computation burdens during training, we incorporate two GCN bottlenecks at the ends of $\Phi$ with a channel reduction ratio of 4. Another residual GCN is adopted within the bottleneck for better information propagation. All GCN layers in our DropGraph share the same constructed $\mathbf{A}$. After average pooling the distortions, following \cite{tang2020beyond}, we apply random multipliers within (0, 1) at all sampled units masked by $\mathbf{M}$.
	
	\textbf{Comparison with state of the art.} DropGraph resembles previous works with probability scheduling and block distortion, but distinguishes them by the novel reformulation of graph reasoning as regularization. Such formulation generalizes other Dropout-based regularizers \cite{tompson2015efficient, ghiasi2018dropblock} as special cases, and is potentially capable of fitting any possible distortion function including Disout \cite{tang2020beyond}. Note that we are not aiming at chasing state-of-the-art performances on all down-streaming tasks. In this work, we implemented the DropGraph in the most basic form consisting of simple convolutions only. A more careful design of the graph network or the propagation rules will lead to substantially better results, whereas we leave this extension to our future work.
	
	Compared to the recent works, R-Drop \cite{wu2021r} and AutoDropout \cite{49960}, our plug-and-play DropGraph is able to function on various architectures universally. R-Drop requires multiple forward pass of the same module during training and adds intermediate supervisions that are hard to balance between the original training objectives; AutoDropout demands highly customized controllers to adjust dropout patterns dynamically for different architectures. Moreover, AutoDropout needs pre-defined search spaces with carefully tuned parameters which imposes much more manual effort and exceeds the run time cost of DropGraph and other counterparts. With close performance, DropGraph yields considerably better accessibility and extensibility during network training. However, due to the lack of resources and open-sourced implementation \footnote{\url{https://github.com/google-research/google-research/issues/727}}, a direct and fair comparison under our experimental setting is impossible.

	\section{Experiments} \label{exp}
	In this section, experimental results are firstly reported on 4 different tasks across 8 datasets. We repeat three independent runs for image classification, semantic segmentation and point cloud classification tasks. For graph recognition task, 100 independent runs were repeated. Subsequently, we conducted extensive studies to analyze DropGraph under different ablative settings with a fixed random seed. All experiments were implemented in PyTorch framework \cite{paszke2019pytorch} using Tesla V100 GPUs. \textbf{More experiments and complete implementation details can be found in the Appendix Sec. \ref{appendix:a} and Sec. \ref{appendix:b}.}
	
		\begin{table}[t]
        \begin{center}
        \setlength\tabcolsep{1.05em}
    	\caption{Comparison results on image classification tasks (\%).}
    		\begin{tabular}{l|cc} 
    			\toprule 
    			CIFAR & CIFAR 10 & CIFAR 100 \\
    			\hline
    			\hline
    			\textcolor{gray}{ResNet-50} & \textcolor{gray}{93.67$\pm$0.11}&\textcolor{gray}{77.73$\pm$0.18}\\
    			+ Dropout &94.81$\pm$0.09 & 77.96$\pm$0.13\\
    			+ SpatialDropout&94.92$\pm$0.11& 78.27$\pm$0.10\\
    			+ DropBlock &95.14$\pm$0.11& 78.71$\pm$0.13\\
    			+ Disout  & 95.25$\pm$0.13& 78.91$\pm$0.15\\
    			+ DropGraph &\textbf{95.35$\pm$0.14}&\textbf{79.58$\pm$0.14}\\
    			\hline
    			\hline
    			\textcolor{gray}{RegNetX-200MF} &\textcolor{gray}{93.10$\pm$0.10}&\textcolor{gray}{71.57$\pm$0.10}\\
    			+ Dropout & 92.77$\pm$0.08 & 71.19$\pm$0.12 \\
    			+ DropBlock & 93.05$\pm$0.13 & 71.21$\pm$0.10 \\
    			+ Disout & 93.15$\pm$0.11 & 71.66$\pm$0.13 \\
    			+ DropGraph&\textbf{93.76$\pm$0.13}&\textbf{71.94$\pm$0.15}\\
    		    \bottomrule
    			\multicolumn{3}{c}{}\\
    			\toprule 
    			ImageNet & Top-1 & Top-5 \\
    			\hline
    			\hline
    			\textcolor{gray}{ResNet-50} & \textcolor{gray}{76.51$\pm$0.07}&\textcolor{gray}{93.20$\pm$0.05}\\
    			+ Dropout &76.80$\pm$0.04& 93.41$\pm$0.04\\
    			+ SpatialDropout&77.41$\pm$0.04& 93.74$\pm$0.02\\
    			+ DropBlock &78.13$\pm$0.05& 94.02$\pm$0.02\\
    			+ Disout  &78.33$\pm$0.06& 93.98$\pm$0.04\\
    			+ DropGraph &\textbf{78.43$\pm$0.04}&\textbf{94.05$\pm$0.04}\\
    			\hline
    			\hline
    			\textcolor{gray}{RegNetX-200MF} &\textcolor{gray}{67.54$\pm$0.08}&\textcolor{gray}{88.29$\pm$0.05}\\
    			+ Dropout & 67.59$\pm$0.06 & 88.28$\pm$0.04 \\
    			+ DropBlock & 67.94$\pm$0.05 & 88.44$\pm$0.04 \\
    			+ Disout & 68.25$\pm$0.06 & 88.56$\pm$0.04\\
    			+ DropGraph &\textbf{68.77$\pm$0.06}&\textbf{88.65$\pm$0.03}\\
    			\bottomrule
    		\end{tabular}
    		\label{img_cls}
    	\end{center}
    	\vspace{-0.5em}
    \end{table}

	\subsection{DropGraph for Image Classification} \label{imgcls}
	
	Two commonly adopted benchmarks were used to evaluate our method on image classification task. CIFAR 10/100 are composed of 60000 images of size $32\times 32$, with the training set containing 50000 images and the validation set containing 10000 images. The images were evenly distributed into 10 and 100 classes. The ImageNet dataset is composed of 1.2M high-resolution training images and 50K validation images, which are distributed into 1K different categories.

	
	To have a comprehensive validation of our DropGraph on both standard and mobile network regimes, we experimented on two network backbones: ResNet-50 \cite{he2016deep} and RegNetX-200MF \cite{radosavovic2020designing}. Only the most \textbf{basic training strategies} were used without advanced tricks such as AutoAugment \cite{cubuk2019autoaugment} or Exponential Moving Average (EMA) of network weights. Note that Disout is originally trained for twice as many epochs as other methods and achieves slightly better results. For fair comparisons, we aligned the training procedures and reproduced all results.
	
	For ResNet-50 backbone, we inserted DropGraph at all building blocks of the last two building groups \cite{tang2020beyond}. For RegNetX-200MF, DropGraph was inserted at the last 3 building blocks of the last building group only. $\alpha$ and $\rho$ are set to $0.2$ and $0.1$ respectively in all related experiments. We directly utilized the block size studied in \cite{ghiasi2018dropblock} with $s=3$ for CIFAR and $s=7$ for ImageNet. 
	
	Table \ref{img_cls} shows the image classification results on the two backbones compared to state-of-the-art regularization methods. Compared to non-regularized baselines, the incorporation of DropGraph brings consistent performance gains on both datasets without affecting network inference. By generating dynamic distortions, our graph-based regularizer learns the best regularization effects on both backbone networks and outperforms both DropBlock and Disout accordingly on all metrics.

	\begin{table}[t]
		\begin{center}
		\caption{Comparison results on semantic segmentation tasks (\%).}
				\begin{tabular}{l|cc} 
					\toprule
					Pascal VOC \ \ \ \ \ \ \ \ \ \ \ \ & \ \ \ \ \ \ mIoU \  \ \ \ \ \ & \ \ \ \ \ \ mAcc \ \ \ \ \ \ \\
					\hline
					\hline

					\textcolor{gray}{FCN-32S}& \textcolor{gray}{47.1$\pm$0.4}& \textcolor{gray}{81.2$\pm$0.3}\\
					+ DropBlock & 50.6$\pm$0.3 & 82.1$\pm$0.2 \\
					+ Disout & 50.7$\pm$0.4 & 82.0$\pm$0.3 \\
					+ DropGraph & \textbf{51.0$\pm$0.4}&\textbf{82.4$\pm$0.2} \\
					\hline
					\hline
					\textcolor{gray}{DeepLabV3}& \textcolor{gray}{53.8$\pm$0.3}& \textcolor{gray}{83.5$\pm$0.3}\\
					+ DropBlock  & 56.9$\pm$0.4 & 83.7$\pm$0.3 \\
					+ Disout & 57.2$\pm$0.4 & 84.0$\pm$0.4 \\
					+ DropGraph & \textbf{58.0$\pm$0.2}& \textbf{84.6$\pm$0.3}\\
					\bottomrule
					\multicolumn{3}{c}{}\\
					\toprule
                    MoNuSeg & mIoU  & DICE \\
					\hline
					\hline
					\textcolor{gray}{U-Net} & \textcolor{gray}{68.2$\pm$0.3}&\textcolor{gray}{80.7$\pm$0.3}\\
				    + DropBlock & 67.6$\pm$0.8 & 80.2$\pm$0.7\\
					+ Disout  & 68.1$\pm$0.8 & 80.0$\pm$0.8\\
					+ DropGraph & \textbf{68.5$\pm$0.7} & \textbf{81.4$\pm$0.6} \\
					\hline
					\hline
                 \textcolor{gray}{Attention U-Net} & \textcolor{gray}{68.3$\pm$0.2}&\textcolor{gray}{81.1$\pm$0.2}\\
					+ DropBlock & 68.0$\pm$0.6 & 81.0$\pm$0.4\\
					+ Disout & 68.3$\pm$0.6 & 81.2$\pm$0.5\\
					+ DropGraph & \textbf{68.7$\pm$0.5} & \textbf{81.5$\pm$0.5}\\
					\bottomrule
				\end{tabular}
			
			\label{voc}
		\end{center}
		\vspace{-1.1em}
	\end{table} 
	
	\subsection{DropGraph for Semantic Segmentation}
	
	DropGraph is then benchmarked on two semantic segmentation datasets: PASCAL VOC 2012 \cite{everingham2015pascal} and MoNuSeg \cite{kumar2017dataset}. The PASCAL VOC benchmark consists of 21 classes with 20 foreground object classes and one background class. There are 1464 training samples and 1449 validation samples in the original dataset. Following \cite{hariharan2011semantic, chen2018encoder}, we used the extra annotations to enriched the dataset to contain 10582 training images. The regularization methods were evaluated on two of the most commonly used segmentation networks, FCN \cite{long2015fully} and DeepLabV3 \cite{chen2018encoder} with \textbf{randomly initialized} ResNet-50 backbone. The quantitative metrics include intersection-over-union (mIoU) and pixel accuracy (mAcc) averaged across the 21 classes. The nuclei segmentation dataset MoNuSeg contains 30 training and 14 testing microscopy images in size $1000^2$. The scans were sampled from different whole slide histopathology images of multiple organs. Following \cite{xiang2020bio}, we enrich the dataset by extracting $512^{2}$ patches at 4 corners. mIoU along with Dice coefficient (DICE) scores are reported for this task. U-Net \cite{ronneberger2015u} and Attention U-Net \cite{oktay2018attention} were adopted as the backbone networks for nuclei segmentation.
    
    For FCN and DeepLabV3, regularizers were applied at the last two building groups of the ResNet-50 backbone and retain the same hyper-parameters used in image classification tasks, such that $\alpha=0.2, \rho=0.1$ and $s=7$. For U-Net based networks, we inserted the regularizers at the last encoder level only with $\alpha=0.1, \rho=0.1$ and $s=7$.

    The segmentation results are presented in Table \ref{voc}. Consistent performance improvements can be observed on both backbone networks and both datasets, hence proving the universal applicability of our DropGraph. Noteworthy, by zeroing out neuron activations, DropBlock leads to even poorer results on many metrics. However, distortion-based methods appeared to be more suitable for such pixel-wise prediction task. Among all the competing methods, DropGraph provides the greatest performance improvements on the randomly initialized backbone networks and shrinks the gap with the ImageNet pre-trained ones.

\begin{table}[t]
    \centering
    \caption{Results on point cloud classification task (\%).}
    \begin{tabular}{c||c c c}
    \toprule
   Methods & \textcolor{gray}{DGCNN} & +Disout & +DropGraph \\
   \hline
   Acc & \textcolor{gray}{92.6$\pm$0.3} & 93.0$\pm$0.3& \textbf{93.2$\pm$0.2} \\
   \bottomrule
    \end{tabular}
    \vspace{-1.1em}
    
    \label{point}
\end{table}

\subsection{DropGraph for Point Cloud Classification}
Apart from studies on image data, we conducted subsequent evaluations of our DropGraph on graph-like data. Since DropBlock is not intrinsically applicable for graph signals, we thus compared DropGraph with Dropout, SpatialDropout and Disout. The ModelNet40 dataset~\cite{modelnet40} was first adopted to verify the potential of DropGraph on 3D point cloud classification task. This 40-class dataset is composed of 9843 training samples and 2468 testing samples. We adopted the same pre-processing instructions as introduced in \cite{pointnet, xiang2021walk} to uniformly sample 1024 points from each raw 3D model. The sampled coordinates were further normalized into unit spheres for better network understanding. We utilized DGCNN \cite{dgcnn} as the backbone network for all experiments. All regularization methods were equipped after each EdgeConv module with $\alpha=0.15, \rho=0.1$ and $s=1$. 
	
	
As demonstrated in Table~\ref{point}, most of the regularizers impact the backbone network positively, except for the plain Dropout that hurts feature representations and leads to inferior results. Among all the competing methods, our DropGraph stands out with the greatest average accuracy improvement (0.6\%) that even outperforms more advanced and complex networks \cite{liu2019relation, yan2020pointasnl}. 
	
	
	\subsection{DropGraph for Graph Recognition}
	
	
	Lastly, DropGraph was evaluated on the semi-supervised node classification task with the Cora dataset \cite{sen2008collective} and on the graph classification task with the Protein dataset \cite{dobson2003distinguishing}. Cora is a citation network dataset, consisting of one graph with 2708 nodes (documents) and 5429 undirected edges (citation links). The nodes are distributed into 7 classes with only 20 labeled nodes per class for training. We followed the same training and testing split as in \cite{kipf2016semi}. The Protein dataset consists of 1113 graphs with an average of 39 nodes and 73 edges per graph. The graphs are binary labeled by their enzymatic activity with no particular testing dataset. We utilized a simple two-layer GCN \cite{kipf2016semi} as the backbone network for both datasets and applied regularizers before the second GCN layer only.  
	
		\begin{table}[t]
        \centering
	\caption{Results on graph recognition tasks (\%).}
		\begin{tabular}{l|cc} 
			\toprule 
			 Methods \ \ \ & Cora & Protein  \\
			\hline
			\hline
			\textcolor{gray}{GCN} & 82.9$\pm$0.5 & 81.5$\pm$0.9 \\
			+ Dropout& 83.1$\pm$0.5 & 80.8$\pm$2.1 \\ 
			+ SpatialDropout & 83.2$\pm$0.6 & 81.6$\pm$1.3 \\
			+ Disout & 82.7$\pm$0.6 &81.7$\pm$1.5 \\
			+ DropGraph & \textbf{83.3$\pm$0.5} & \textbf{82.1$\pm$1.7} \\
			\bottomrule
			\multicolumn{3}{c}{}\\
			\toprule 
			GCN & ppa Val & ppa Test\\
			\hline
			\hline
			\textcolor{gray}{+ Dropout} & 65.0$\pm$0.0 & 68.4$\pm$0.0 \\ 
			+ DropGraph & \textbf{69.9$\pm$0.0} & \textbf{73.2$\pm$0.0} \\
			\bottomrule
			\multicolumn{3}{c}{}\\
			\toprule 
			Cluster-GCN & products Val & products Test\\
			\hline
			\hline
			\textcolor{gray}{+ Dropout} & 92.1$\pm$0.0 & 79.0$\pm$0.0 \\ 
			+ DropGraph & \textbf{92.9$\pm$0.0} & \textbf{80.2$\pm$0.0} \\
			\bottomrule
		\end{tabular}
		\vspace{-0.5em}
	\label{pr}
\end{table}
	
	We further examined our proposed method on two large scale datatsets: opga-ppa and ogbn-products from the Open Graph Benchmark (OGB) \cite{hu2020open}. We followed the official implementations of the backbone networks and inserted our DropGraph between every pair of GCN layers. In these experiments, $\alpha$ is set to $0.15$ for all the tasks while $\rho$ is set to $0.1$ for Cora, OGB datasets, and $0.25$ for Protein.
	 
	 The results reported in Table \ref{pr} demonstrate that among all competing methods, DropGraph yields the greatest average accuracy gains with $0.4\%$ on the Cora dataset and $0.6\%$ on the Protein dataset. DropGraph also achieves superior performances on both validation and testing set of the two large-scale OGB benchmarks.
	
	\subsection{Ablation Studies and Analysis} \label{ablation}
	Unless explicitly mentioned, we adopted ResNet-50 as the backbone network and conducted ablative experiments on ImageNet dataset with the same training configurations specified in Sec. \ref{imgcls}. The linear distortion probability scheduler \cite{ghiasi2018dropblock} was used for all ablation studies. 


 
	\textbf{Studies on different designs of $\mathbf{A}$.} Different dependency modeling strategies may impact the regularization effect to different extents. 
	To benchmark our design in Eq. \ref{eq5}, we studied four alternatives of $\mathbf{A}$ that models distinct intra- and inter-dependencies: \textbf{(a)} $\mathbf{A}$ is learned jointly with backbone network and hence is independent of $\mathbf{V}$. \textbf{(b)} Similar vertices are assigned with high values, such that $\mathbf{A}=\texttt{softmax}(\texttt{sim}(\mathbf{V},\mathbf{V}))$. \textbf{(c)} Only intra-dependencies are modeled, such that $\mathbf{A}=\mathbf{I}$ becomes an identity matrix. \textbf{(d)} Both intra-dependencies and inter-dependencies are equally modeled with $\mathbf{A}=\frac{\mathbbm{1}}{||\mathbf{V}||}$.
	
	\begin{table}[h]
    	\begin{center}
    	\caption{Comparisons on designs of the adjacency matrix $\mathbf{A}$.}
    		\begin{tabular}{c|cc} 
    			\toprule 
    			\ \ \ Designs of $\mathbf{A}$  \ \ \ & \ \ \ Top-1 (\%) \ \ \ & \ \ \ Top-5 (\%) \ \ \ \\
    			\hline
    			\hline
    			\textcolor{gray}{Backbone} &\textcolor{gray}{76.51}&\textcolor{gray}{93.20}\\
    			\hline
    			\textbf{(a)} & 77.74 & 93.68 \\
    			\textbf{(b)} & 78.13 & 93.88 \\
    			\textbf{(c)} & 78.02 & 93.89 \\
    			\textbf{(d)} & 78.25 & 93.98 \\
    			\hline
    			Eq. \ref{eq5}& \textbf{78.40} & \textbf{94.07} \\
    			\bottomrule
    		\end{tabular}
    		\label{ablation_a}
    	\end{center}
    	\vspace{-0.5em}
    \end{table}

	The ablative results of the four designs are numerically compared in Table \ref{ablation_a}. DropGraph brings consistent performance gains to the backbone network with all adjacency matrix designs. We observed from experiment \textbf{(a)} that learning a static $\mathbf{A}$ that is insensitive to the incoming feature distributions hurts the overall regularization effect provided by DropGraph. Inferior performances are also obtained when modeling stronger dependencies for similar vertices, as indicated by experiments \textbf{(b)} and \textbf{(c)}. Surprisingly, experiment \textbf{(d)} reveals that constructing equal connections between all pairs of vertices also leads to effective regularization and ends up with even higher results than DropBlock.
	
	In our hypothesis, dissimilar features in $\mathbf{V}$ should be strongly correlated to weaken the locally similar ones. Therefore, we expect to see the strong modeling of similar features \textbf{(b)} and identity features \textbf{(c)} to be inferior to a weak modeling of similar features \textbf{(d)} and strong modeling of dissimilar features (Eq. \ref{eq5}). The experimental results agreed with our hypothesis.
	
	
    
	

    \textbf{Studies on $\alpha$ and $\rho$.} DropGraph is controlled by the sampling ratio $\alpha$ and the distortion probability $\rho$. A higher $\alpha$ provides a more complete feature distribution and a higher $\rho$ imposes stronger regularizations. 
    
\begin{figure}[t]
		\centering
		\includegraphics[width=\linewidth]{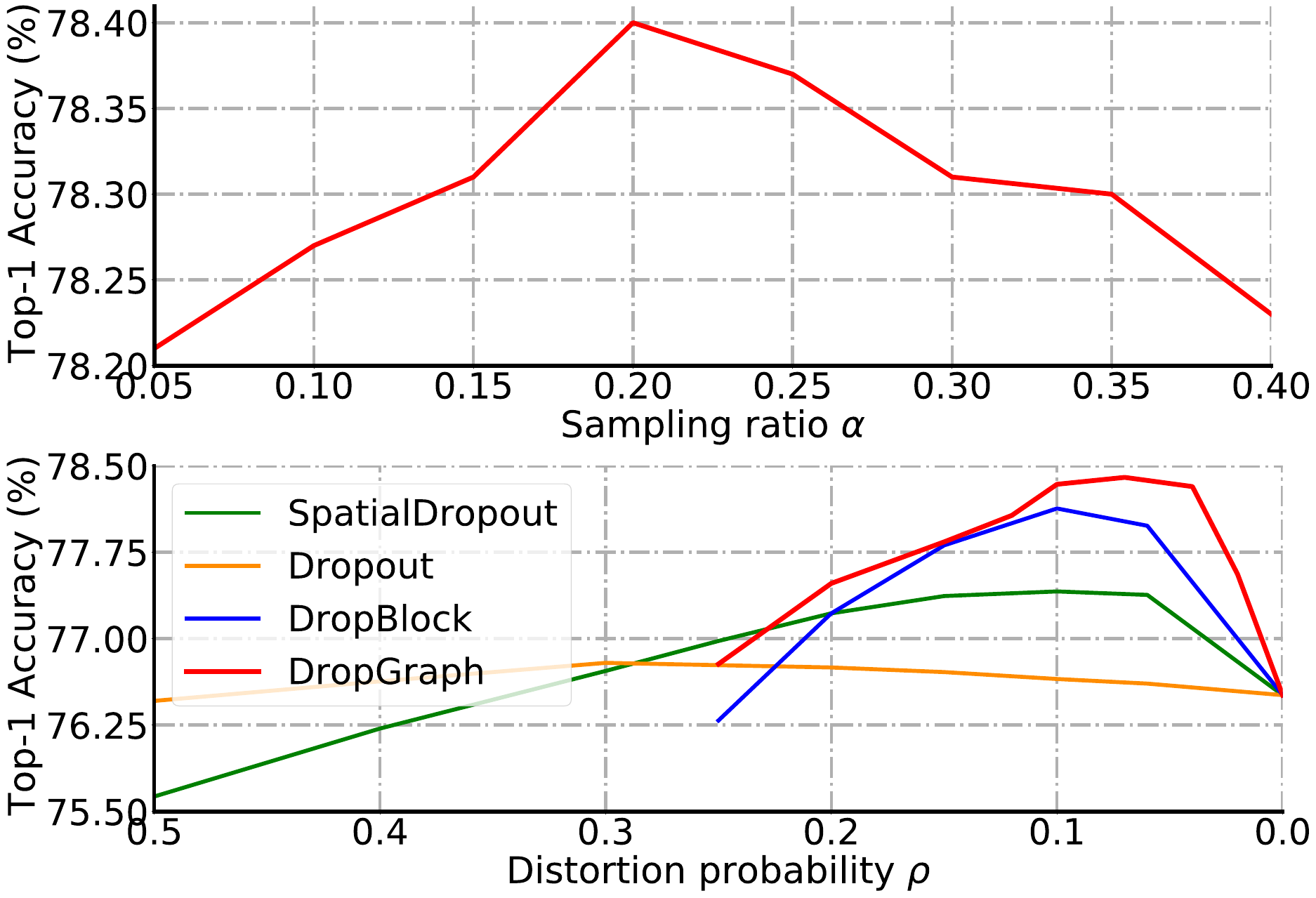}
		\caption{\textbf{Top:} Sampling rate $\alpha$ \textit{v.s.} ImageNet accuracy. \textbf{Bottom:} Distortion probability $\rho$ \textit{v.s.} ImageNet accuracy. Comparison results are collected from \cite{ghiasi2018dropblock}.}
		\label{fig:ablation}
		\vspace{-0.5em}
	\end{figure}

    In Figure \ref{fig:ablation} top, we showed that different settings of $\alpha$ bring consistent accuracy improvement to the backbone network that universally outperforms DropBlock. When $\alpha=0.2$, DropGraph reaches the highest top-1 accuracy of $78.4\%$. Our hypothesis is two-fold: \textit{(i)} After softmax normalization (Eq. 6), $\mathbf{A}$ with more samples will have much closer one-to-all similarity. In this way, dependencies between any two samples will become hardly discriminated when the sample number increases. Poorer dependency modeling will directly impact the overall performance, as also pointed out in Table 4. \textit{(ii)} More samples require better reasoning ability of the graph network. Since our graph network is constructed with only few basic GCN layers, it cannot guarantee effective distortions for large numbers of samples at a time. The experiments also validated that the sampling rate should be not too small nor too big, and 0.2 sampling rate yields the best performance.
    
    With the best $\alpha$ found, in Figure \ref{fig:ablation} bottom, we demonstrated that with distortion (drop) probability $\in (0.0, 0.2)$, DropGraph significantly outperforms the simplest Dropout and SpatialDropout. Under the same set of $\rho$ values studied in \cite{ghiasi2018dropblock}, DropGraph consistently surpasses the DropBlock baseline with identical inference behaviour.
	


	
\textbf{Grad-CAM masking classification.} Grad-CAM \cite{selvaraju2017grad} is widely used to visualize the focus of networks. We are curious whether the focus of backbone network would change when applying different regularization methods. Therefore, we here conduct a set of quantitative experiments to inspect the network focus using the CIFAR 10 dataset, which were set up as follows: \textit{(i)} Grad-CAM heatmaps were generated using the regularized backbones. \textit{(ii)} The heatmaps were binarized with a 50\% threshold for each image instance. \textit{(iii)} The original images were then masked by the thresholded binary heatmaps and we trained an additional ResNet-56 on the masked RGB images. 

\begin{table}[h]
    \centering
    \caption{Grad-CAM heatmaps masked CIFAR-10 classification. }
    \begin{tabular}{c||c c c}
    \toprule
   Methods & +Dropout & +Disout & +DropGraph \\
   \hline
   Acc (\%) & 51.25$\pm$0.22 & 58.69$\pm$0.21 & \textbf{65.64$\pm$0.17} \\
   \bottomrule
    \end{tabular}
    
    \label{grad_cam}
\end{table}

In Table \ref{grad_cam}, we found that masking the images with our DropGraph yields the highest classification accuracy, thus validating the heatmaps indeed focus on the primary objects better. 

\noindent
\textbf{Distortion generation methods.} As mentioned in Sec. \ref{dropgraph}, we hypothesize that GCN is able to aggregate long range uncorrelated signals with specific underlying patterns. However, other perturbation methods, such as standard convolution, feature average pooling, and noise injection, either cannot model meaningful long range dependencies or yield no capturable patterns for the backbone network. To validate our hypothesis empirically and claim the necessity of using the stand-alone graph network to generate distortions, we examined a series of different methods in Table \ref{methods}. 

	\begin{table}
    \centering
	\caption{Comparisons on different distortion generation methods. Graph represents the proposed usage of GCN.}
		\begin{tabular}{ l | c }
			\toprule
			Methods & CIFAR-100 (\%)\\
			\hline 
			\hline
			\textcolor{gray}{Backbone}& \textcolor{gray}{77.73$\pm$0.18} \\
			+ random noise & 78.71$\pm$0.14\\
			+ avg pooling & 78.88$\pm$0.11\\
			+ standard conv & 78.41$\pm$0.10\\
			+ dilated conv (r=3) & 78.74$\pm$0.13\\
			+ dilated conv (r=3) \& avg pooling & 79.09$\pm$0.12\\
			\hline
			+ GCN \& avg pooling (ours) & 79.44$\pm$0.13\\
			+ GCN (ours) & \textbf{79.58$\pm$0.14}\\
			\bottomrule
		\end{tabular}
		\label{methods}
	\end{table}

Our conclusion from this experiment is two-fold: \textit{(i)} Generating distortions with short range features provides weak regularization effects. This aligns with our claim in Sec. \ref{gr}. \textit{(ii)} For dilated conv generated distortions, using the one average pooled distortion to override all original features leads to slightly better results, whereas using GCN-based generation method to generate different distortions for different samples is able to achieve the best results. Therefore, such empirical study validates that our proposed DropGraph framework along with the partial graph reasoning strategy is the most effective method among all other possible designs.

\noindent\textbf{Limitations.} With the additional graph, DropGraph yields a bit more complex framework than the DropBlock baseline: \textit{(i)} Same as Disout, DropGraph requires three hyper-parameters ($\alpha, \rho, s$) which are one more than DropBlock. However, in Sec. \ref{ablation}, we demonstrate that DropGraph generally surpasses DropBlock without explicit tuning of the hyper-parameters; \textit{(ii)} DropGraph puts additional computations to the backbone network during training. Nevertheless, the training computation overhead is trivial that DropGraph only imposes less than $1\times10^{-3}$ G extra MACs on the ResNet-50 backbone and merely 4\% more training time compared to Disout.

	\section{Conclusion} \label{conclusion}
	In this paper, we introduce the very first learning-based framework for neural network regularization, namely DropGraph. Instead of zeroing out information as in Dropout, we build a stand-alone graph neural network to generate feature map distortions. DropGraph provides Dropout-based regularization by sampling graph nodes from partial feature maps with dependencies modeled via the adjacency matrix and distortions generated through a set of graph reasoning layers. Our DropGraph outperforms other state-of-the-art regularizers on a variety of tasks and datasets.


\bibliography{example_paper}
\bibliographystyle{icml2022}

\clearpage
\appendix
\section{Extensive Experiments on Scheduler} \label{appendix:a}

\begin{figure}[h]
\centering
\hspace{-3em}
\begin{minipage}[]{\linewidth}
\centering
\includegraphics[width=\linewidth]{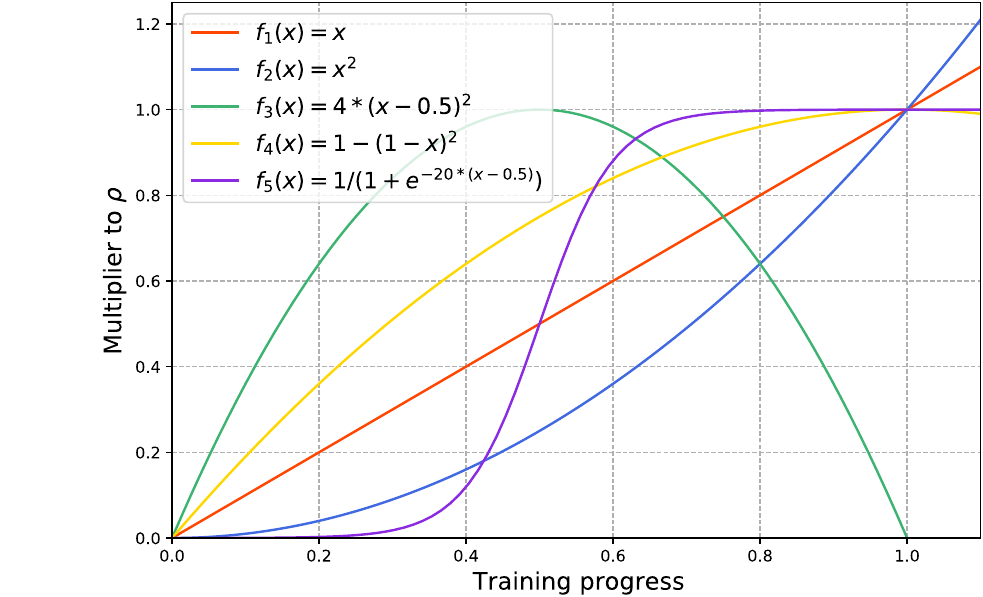}
\end{minipage}%
\vfill
\centering
\begin{minipage}[c]{\linewidth}
\centering
\begin{tabular}{c|cc} 
		\toprule 
		\ \ \ Scheduler \ \ \ & \ \ \ Top-1 (\%) \ \ \ & \ \ \ Top-5 (\%) \ \ \ \\
		\hline
		\hline
		$f_1$ & 78.40 & \textbf{94.07} \\
		$f_2$ & 77.36 & 93.58 \\
		$f_3$ & 77.94 & 93.85 \\
		$f_4$ & 77.86 & 93.83 \\
		$f_5$ & \textbf{78.43} & 94.05 \\
		\bottomrule
	\end{tabular}

\end{minipage}
\caption{\textbf{Top} Five candidate schedulers that adjust the distortion probability from $0$ to $\rho$. \textbf{Bottom:} ImageNet validation results on the five candidate schedulers.} \label{scheduler}
\end{figure}

As mentioned in \cite{ghiasi2018dropblock,zoph2018learning}, increasing $\rho$ while training proceeds helps network understanding at the initial learning stage. We here provide an additional study on 5 different $\rho$ scheduling functions that are shown in Figure \ref{scheduler} top. According to the quantitative results reported in Figure \ref{scheduler} bottom, using $f_2$ leads to the poorest results. This is because $f_2$ applies the weakest regularization effect as it is the only scheduler that yields the lowest distortion probabilities at all training steps. On the other hand, DropGraph achieves the best top-1 accuracy with scheduler $f_5$ and the best top-5 accuracy with the linear scheduler. Note that with the same linear scheduler, DropGraph already outperforms DropBlock and Disout by a considerable margin.  

\section{Experimental Details} \label{appendix:b}

\subsection{Detailed Settings for Image Classification}

\noindent
\textbf{CIFAR.} We used the open-source PyTorch implementation \footnote{https://github.com/kuangliu/pytorch-cifar} for all experiments on the CIFAR datasets. SGD with weight decay of 0.0005 and momentum of 0.9 was used as the optimizer. The batch size was set to 128 for both ResNet-50 and RegNetX-200MF backbones. The learning rate initially began at 0.1 and reduced by a factor of 0.1 at epochs 150 and 250 while all models were trained for 300 epochs from scratch. Standard augmentation techniques were adopted during training, including random cropping of a $32\times 32$ sample from the 4-pixel padded images and random horizontal flipping with a probability of 0.5. All images were normalized by their mean and standard deviation in the pre-processing step.

\noindent
\textbf{ImageNet.} For fair comparisons, we borrowed the implementations provided by \cite{tang2020beyond} \footnote{https://github.com/huawei-noah/Disout} and slightly change the total training epochs and optimizer scheduler to align with \cite{ghiasi2018dropblock}. Specifically, we used SGD optimizer with weight decay of 0.0001 and momentum of 0.9 for optimizations. The batch size was set to 1024 for all experiments that were equally distributed on 8 V100 GPUs. All models were trained for 270 epochs starting with an initial learning rate of 0.1 and reduced by 0.1 at epoch 125, 200 and 250 without learning rate warming up. During training, input images are augmented by cropping into $224\times224$ patches and randomly horizontal flip. During validation, images are first scaled into $256\times256$ and then center cropped into $224\times224$ before being fed into the networks. The above training configurations are consistent to the ones used in \cite{he2016deep, ghiasi2018dropblock, tang2020beyond}.

\subsection{Detailed Settings for Semantic Segmentation}
\noindent
\textbf{Pascal VOC 2012.} The experiments were conducted in a public framework \footnote{https://github.com/warmspringwinds/pytorch-segmentation-detection}. We used the Adam optimizer \cite{kingma2014adam} with 0.0001 weight decay to minimize the cross entropy loss. Learning rate starts from 0.001 and cosinanealing scheduled to 1$e^{-5}$ in 400 epochs. The batch size was set to 64 for both backbones. Training images are randomly horizontal flipped, randomly scaled by a factor within [0.5, 1.5] and then center cropped leaving with $224\times 224$ patches to fit the ResNet-50 backbone. Note that all backbone networks were trained from scratch rather than being pre-trained on ImageNet as suggested in \cite{ghiasi2018dropblock}.

\noindent
\textbf{MoNuSeg.} Implementations from \cite{xiang2020bio} \footnote{https://github.com/tiangexiang/BiO-Net} were adopted for this experiment. Cross entropy loss was minimized by the Adam optimizer with an initial learning rate of 0.01 and a decay rate of 0.00003 at per step. The training dataset was augmented through random rotation (with the angles from [-15, +15]), random x-y shifting (with the angles from [-5\%, 5\%]), random shearing, random zooming (within [0, 0.2]), and random flipping (both horizontally and vertically). The batch size was set to 2.

	\begin{figure*}[t]
		\centering
		\includegraphics[width=0.95\linewidth]{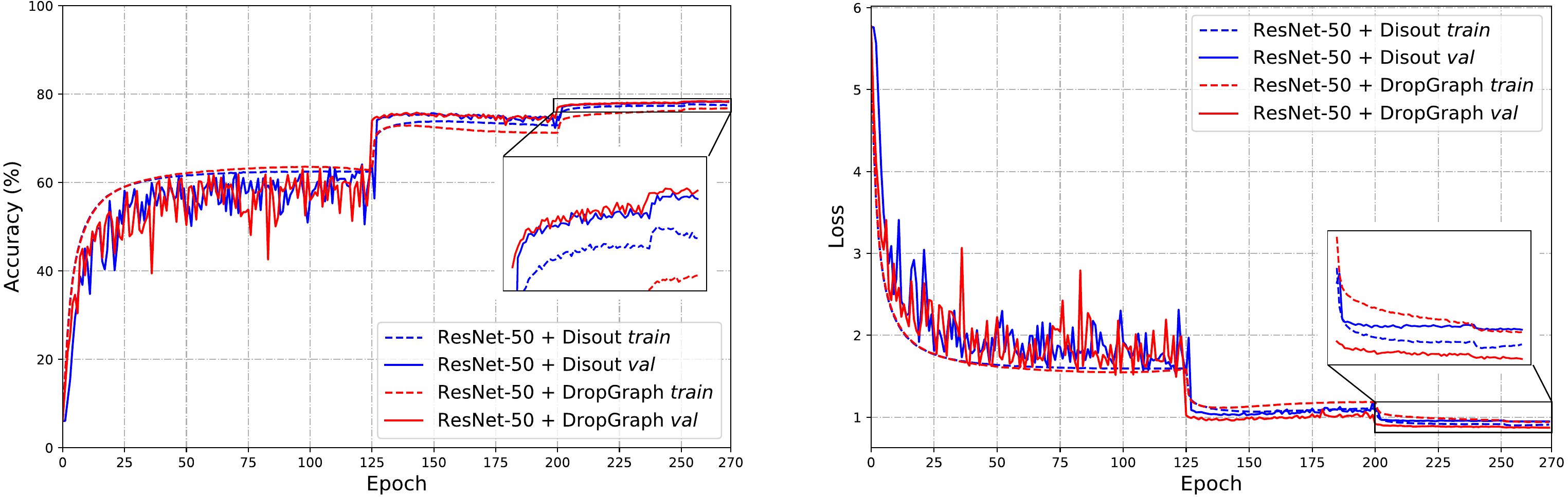}
		\caption{\textbf{Left:} Accuracy curves. \textbf{Right:} Loss curves. Results at the last 70 epochs are zoomed in.}
		\label{fig:curves}
	\end{figure*} 
	
	\begin{figure*}[t]
	\centering
	\includegraphics[width=\linewidth]{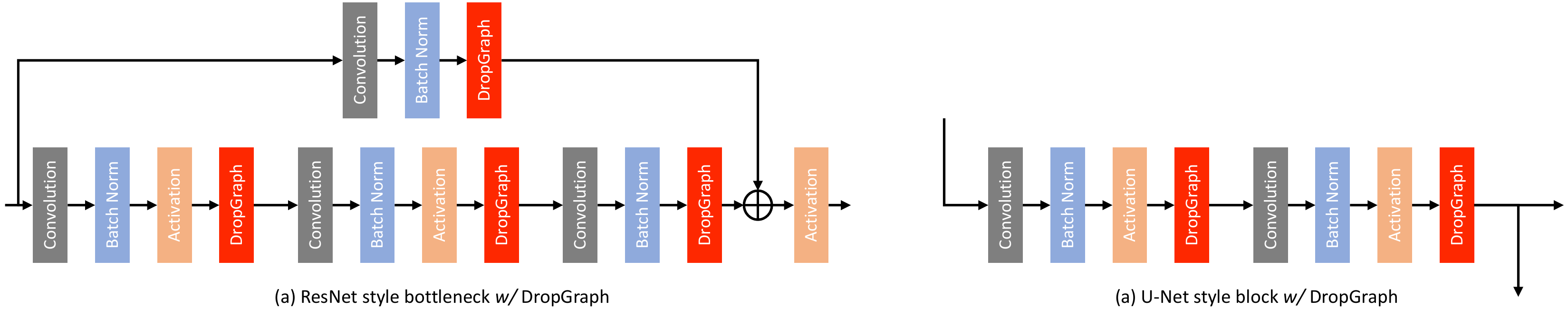}
	\caption{\textbf{Deployment position of DropGraph on ResNet and U-Net.}}
	\label{fig:place}
\end{figure*} 

\subsection{Detailed Settings for Point Cloud Analysis}
We used the open-source implementation \footnote{https://github.com/WangYueFt/dgcnn} for the point cloud classification experiments. We set the height and width of the 2D CNN feature maps of DropGraph and Disout to be equal to the number of points. SGD optimizer with weight decay of 0.0001 and momentum of 0.9 were used for the optimization. We utilized the cosineanealing scheduler to adjust the learning rate from 0.1 to 0.0001 in 250 epochs. The batch size was set to 32 for training and 16 for validation. Raw point clouds were first normalized into unit spheres followed by random scaling with a multiplier within [0.66, 1.5], random translation along the three directions by displacements within [-0.2, 0.2]. The number of neighbors in KNN was set to 20 for all experiments.

\subsection{Detailed Settings for Graph Recognition}

\noindent
\textbf{Cora.} The official GCN implementation \footnote{https://github.com/tkipf/pygcn} was used for experiments on the Cora dataset. Adam optimizer with learning rate of 0.01 and weight decay of 0.0005 was used for optimization. For each run, we trained the models for 200 epochs without batch processing. 

\noindent
\textbf{Protein.}
We used the open-source implementation \footnote{https://github.com/cszhangzhen/HGP-SL} for experiments on the Protein dataset with the same training and validation split suggested in \cite{zhang2019hierarchical}. The optimizer and network were set identically to the ones used for Cora experiments, except that we placed an additional linear layer at the end of the network. Learning rate was set to 0.02 for training on the entire dataset for 200 epochs.

\noindent
\textbf{Open Graph Benchmark.} We used the exact official settings \footnote{https://github.com/snap-stanford/ogb}, including dataloaders, network implementations, and optimizations without any modifications. 

	\begin{figure*}[t]
		\centering
		\includegraphics[width=\linewidth]{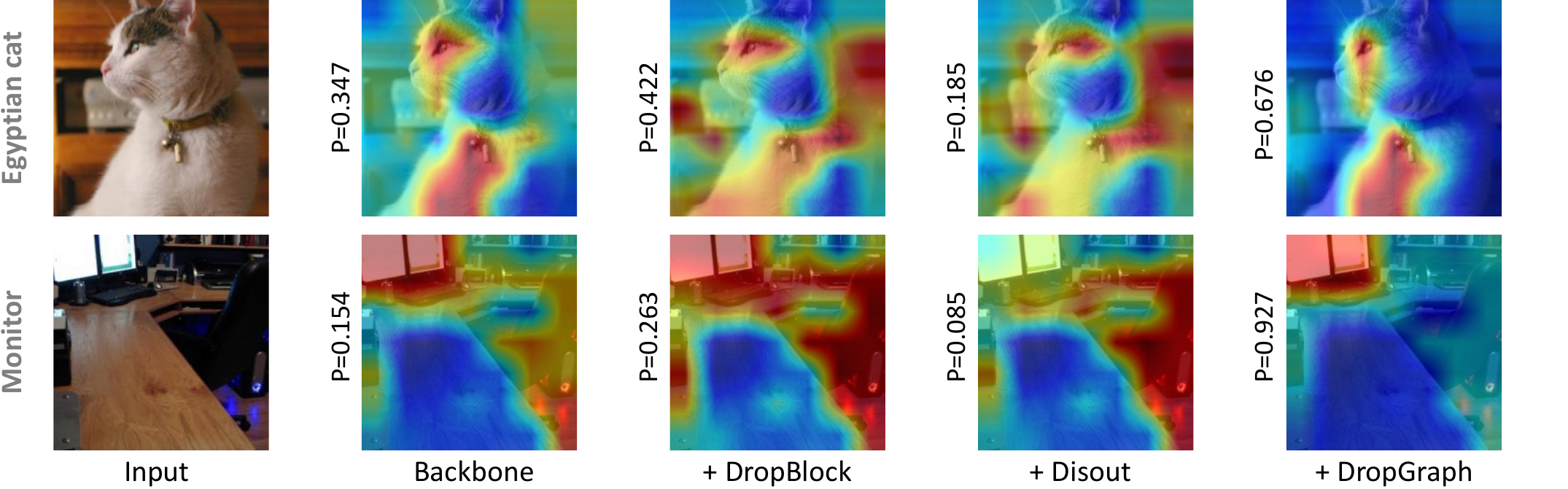}
		\caption{\textbf{Grad-cam visualizations.} Two randomly selected cases from the ImageNet validation set are shown. Gradients are collected toward their predicted classes.}
		\label{fig:gradcam}
	\end{figure*} 
	
\begin{figure*}[t]
		\centering
		\includegraphics[width=\linewidth]{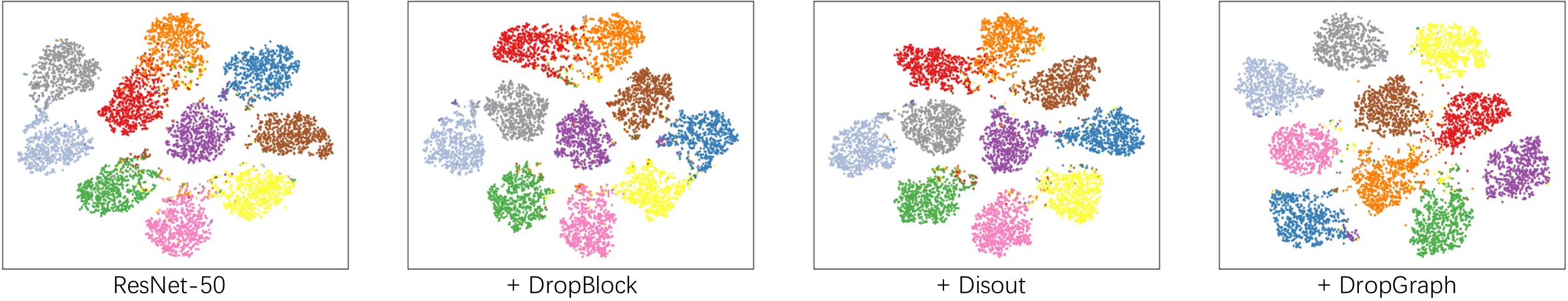}
		\caption{\textbf{t-SNE visualizations on feature representations.} Different classes are marked in different colors.}
		\label{fig:tsne}
\end{figure*}

\section{ImageNet Training Curves}
Here we compare the accuracy and loss curves between DropGraph and Disout during ImageNet training in Figure \ref{fig:curves}. Validation and training statistics are plotted in solid lines and dotted lines respectively, with red denote our DropGraph and blue Disout.

An effective regularizer is able to alleviate over-fitting on the training set and improve the generalization ability on the validation set. According to Figure \ref{fig:curves} left, DropGraph yields generally lower training accuracy and higher validation accuracy. As shown in Figure \ref{fig:curves} right, DropGraph yields higher training loss with lower validation loss. Therefore, our DropGraph demonstrates stronger regularization effects than Disout.

Noteworthy, without modifying the backbone network, all existing regularization methods yield very similar learning curves. Our proposed method is the very first learnable regularizer in the community. Without any advanced designs, the most basic DropGraph implementation already achieves on par and even better results than all existing state-of-the-arts.

\section{Deployment Positions of DropGraph}
In Figure \ref{fig:place}, we show where to apply DropGraph on ResNet and U-Net style networks. As mentioned in Sec. \ref{dropgraph} and Sec. \ref{exp}, in general, we apply DropGraph after each activation layer. Note that we also apply distortions to skip features in all networks with residual connections. 

\section{Visualizations}

\subsection{Grad-CAM Visualizations}

Additional to the quantitative experiments conducted in Sec. \ref{ablation}, we visualize the Grad-CAM heatmaps here for more intuitive demonstrations. The gradients flowed through last convolutional layer are collected for visualization. Figure \ref{fig:gradcam} compares the heatmaps on two input images with different regularizers. Without regularization, backbone network can be easily distracted and is therefore hard to focus on the primary object. The same problem still exists when employing DropBlock and Disout in the backbone network. However, our DropGraph greatly alleviates such distraction and forces the backbone network to apply more attention to the major object features.

\subsection{t-SNE of Feature Representations}

Regularization methods enable the backbone network to better discriminate input samples. We here adopt t-SNE \cite{van2008visualizing} to visualize the feature representations in CIFAR 10 validation set in Figure \ref{fig:tsne}. The representations are collected before the last linear layer of the ResNet-50 model. Compared to DropBlock and Disout, the features generated by DropGraph regularized network have less false predictions and more centered clustered based on their GT labels. Hence, it is validated that our method can achieve better feature representation via a more explainable manner.

\subsection{Qualitative Semantic Segmentation Results}

\begin{figure*}[h]
	\centering
	\includegraphics[width=\linewidth]{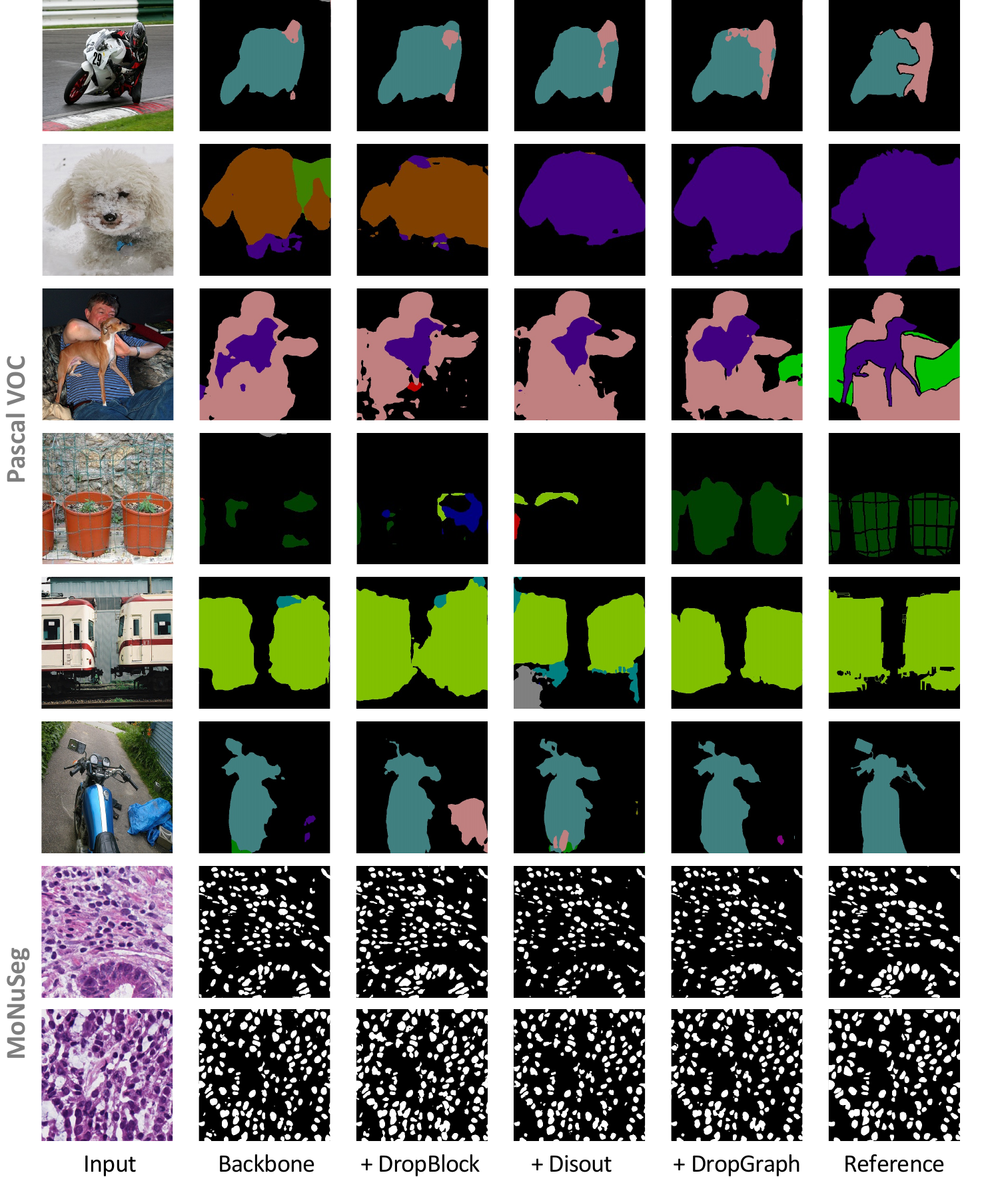}
	\caption{\textbf{Qualitative results on semantic segmentation.}}
	\label{fig:seg_res}
\end{figure*}

We present several qualitative segmentation results in Figure \ref{fig:seg_res}  to intuitively demonstrate the impacts of different regularization methods to the backbone networks. With the network architecture unchanged and identical inference behaviour, DropGraph regularized network infers better segmentation masks that are closer to the references.

\end{document}